\documentclass{article}

\PassOptionsToPackage{numbers, compress}{natbib}
\usepackage[preprint]{neurips_2025}

\usepackage[utf8]{inputenc}
\usepackage[T1]{fontenc}
\usepackage{hyperref}
\usepackage{url}
\usepackage{booktabs}
\usepackage{amsfonts}
\usepackage{amsmath}
\usepackage{nicefrac}
\usepackage{microtype}
\usepackage{xcolor}
\usepackage{enumitem}
\usepackage{tcolorbox}
\usepackage{placeins}
\usepackage{multirow}
\usepackage{caption}
\usepackage{scrextend}

\title{Internalized Reasoning for Long-Context Visual Document Understanding}

\author{%
Austin Veselka \\
LightOn \\
\texttt{austin.veselka@lighton.ai}
}

\begin{document}
\raggedbottom

\maketitle

\begin{abstract}

Visual long-document understanding is critical for enterprise, legal, and scientific applications, yet the best performing open recipes have not explored reasoning, a capability which has driven leaps in math and code performance.
We introduce a synthetic data pipeline for reasoning in long-document understanding that generates thinking traces by scoring each page for question relevance, extracting textual evidence and ordering it from most to least relevant. We apply SFT to the resulting traces within \texttt{<think>} tags, gated by a \texttt{<cot>} control token, and the resulting reasoning capability is internalized via low-strength model merging.
We study Qwen3 VL 32B and Mistral Small 3.1 24B. With Qwen3 VL, we achieve 58.3 on MMLongBenchDoc, surpassing the 7$\times$ larger Qwen3 VL 235B A22B (57.0). With Mistral, we show that synthetic reasoning outperforms distillation from the Thinking version's traces by 3.8 points on MMLBD-C, and internalized reasoning exhibits 12.4$\times$ fewer mean output tokens compared to explicit reasoning.
We release our pipeline for reproducibility and further exploration\footnote{\url{https://github.com/lightonai/distilabel/tree/lc_sft_pipelines}. Built on a fork of an Apache 2.0 licensed package.}.

\end{abstract}

\section{Introduction}
\label{sec:intro}

Long-context (LC) visual language models (VLMs) for document understanding have natural applications in enterprise, legal, scientific and financial domains. The promise of powerful LC VLMs is an engine capable of performing a multitude of complex operations over documents in these domains: specific information retrieval, in-depth understanding for question-answering (QA), complex and multi-hop reasoning and QA, summarization and more. 

Reasoning models \citep{o1} have introduced test-time scaling as a new dimension of scaling for LLMs, with arbitrary length inference sequences removing bounds on the computable algorithms for transformers \citep{rasp,expressive_cot,cot_serial} and implying the potential to scale performance generally, including for the long-document operations above. While reasoning has proven extremely effective in domains with verifiable rewards, e.g. math and coding \citep{o1,deepseek_r1,gpt53_codex}, the lack of such high-signal rewards in QA limits most existing works to non-reasoning approaches. For instance, Docopilot \citep{docopilot} is an open VLM which builds a dataset from Arxiv, Sci-Hub and OpenReview and fine-tunes InternVL 2 2B and 8B with SFT. Recent work \citep{orion} has explored recipes for training long-document VLMs using CPT, SFT and LongPO, but does not explore methods for or benefits of reasoning over the document.


The value of this domain has spurred the creation of challenging benchmarks such as MMLongBenchDoc (MMLBD) \citep{mmlbd} which evaluate the model on documents up to hundreds of pages long with human written questions and answers. Only recently has any model surpassed GPT 4o \citep{gpt4o} in performance (beginning with Qwen3 VL \citep{qwen3vl}) and Qwen3 VL 235B A22B still falls well short of the human baseline, making it clear that this domain is still significantly under-explored. Additionally, Qwen3 VL 235B A22B Instruct and Thinking achieve equivalent performance, highlighting the discussed gap in leading models: frontier reasoning capabilities do not contribute to stronger document understanding. 


Motivated by this gap, we propose a synthetic data pipeline that generates structured reasoning traces via page-level evidence extraction, relevance scoring and bounded top-$K$ sorting. We train Qwen3 VL 32B Instruct\footnote{Released under the Apache 2.0 license. \label{apache_2.0}} (Qwen3 VL) and Mistral Small 3.1 24B\footref{apache_2.0} (Mistral) \citep{mistral31small} and show that low-strength model merging via task arithmetic \citep{task_arithmetic} makes the reasoning implicit. 

Further, there is a background of work related to internalized or implicit chain-of-thought reasoning. These methods include auxiliary losses over the answer rationale \citep{distill_step_by_step,dual_head_reasoning_distillation}, reasoning post answer \citep{cot_after_mysteries,cot_after_adaptive_thinking} and hidden state distillation from thinking models \citep{implicit_cot}, but these works do not clarify whether the rationale improves training signal or modifies inference behavior. We believe our work is the first to show an implicit CoT that is an active inference-time capability rather than a training artifact alone.

\paragraph{Contributions.}
\begin{itemize}[leftmargin=1.2em]
    \item \textbf{Synthetic reasoning pipeline.} We present a synthetic reasoning pipeline for long-document VQA which achieves SOTA on MMLBD and MMLBD-C \citep{orion} with Qwen3 VL 32B Instruct, surpassing the teacher model Qwen3 VL 235B A22B with 7× fewer parameters. We further show that the method generalizes across model families with Mistral improving by +19\% relative to the base model on both MMLBD and MMLBD-C.
    \item \textbf{Internalized reasoning.} We show that reasoning can be \textit{internalized} by low-strength model merging, exhibiting similar output lengths without thinking tokens to the no-think baseline (250 vs 140 for Qwen, 132 vs 160 for Mistral) relative to higher merge strength (Mistral $\alpha = 0.5$) which uses thinking tokens in 77\% of questions and averages 1637 output tokens. 
    \item \textbf{Internalized reasoning is an inference-time capability.} By gating reasoning behind a control token during training, we make it possible to study the effect of the reasoning \textit{during inference}. We use this to show that our internalized reasoning is an inference-time capability with a significant impact on performance: removing the \texttt{<cot>} control token degrades performance on MMLBD by 3.8 points for Qwen and 2.1 points for Mistral.
    \item \textbf{Synthetic reasoning outperforms thinking traces.} We find that our synthetic reasoning pipeline is more effective than traces from Qwen3 VL 235B A22B Thinking, indicating the strength of our trace design and highlighting the potential for future models to improve reasoning for long-document VQA.
\end{itemize}

\section{Related work}

\paragraph{Long-context VLMs and training.}
Most prior work on long-context VLMs focuses on video processing~\citep{longvila,longvita,bolt}, while long-document understanding has been addressed primarily by closed models (GPT-4o~\citep{gpt4o}, Gemini~\citep{gemini25}) or open-weight models with underspecified training recipes~\citep{qwen3vl,glm4v}. Docopilot~\citep{docopilot} fine-tunes InternVL2 on a curated 758K-sample document dataset and V2PE~\citep{v2pe} extends VLM context to 1M tokens via variable position encoding. On the training methodology side, ProLong~\citep{prolong} establishes data recipes for efficient long-context language model training, while LongPO~\citep{longpo} and SoLoPO~\citep{solopo} use short-to-long preference optimization. Synthetic data generation for long-context training includes hard-negative construction~\citep{nextlong}, bootstrapping from short-context models~\citep{bootstrap}, and realistic instruction synthesis~\citep{wildlong}. In \citep{orion}, the authors provide open CPT, SFT, and LongPO recipes for document VLMs and achieve SOTA on MMLBD-C. None of the above works explore the benefits of reasoning over the document context. As such, we extend the recursive answer generation pipeline from \citep{orion} to the task of training reasoning models for long-document VQA. 

\paragraph{Reasoning models.}
Chain-of-thought prompting~\citep{cot} demonstrated that step-by-step reasoning substantially improves LLM performance, spurring a paradigm shift toward trained reasoning models. OpenAI o1~\citep{o1} and DeepSeek-R1~\citep{deepseek_r1} learn extended reasoning via reinforcement learning and thinking variants of Qwen3 VL~\citep{qwen3vl} and GLM~\citep{glm4v} extend this paradigm to multimodal models. LongFaith~\citep{longfaith} generates faithful long-context reasoning traces with citation grounding and Temporal CoT~\citep{temporalcot} applies frame-level evidence selection for long videos---analogous to our page-level approach, but executes the pipeline during inference. Reasoning advances have been concentrated in mathematical and coding benchmarks; for long-document understanding, despite curriculum-guided RL methods such as QwenLong-L1~\citep{qwenlong_l1}, the thinking paradigm shows minimal benefit for Qwen3 VL. 

\paragraph{Implicit chain-of-thought reasoning.}
A growing body of work shows that chain-of-thought supervision at training time can improve test-time performance without requiring reasoning generation during inference. We summarize a few main methodologies in this area. \citep{distill_step_by_step,dual_head_reasoning_distillation} utilize an auxiliary objective over rationales and a train-only reasoning head respectively. \citep{cot_after_mysteries,cot_after_adaptive_thinking} append reasoning after answers. An alternative line distills reasoning into hidden-state or latent representations via objectives such as REINFORCE \citep{quiet_star}, knowledge distillation \citep{implicit_cot} and VAE inspired objectives \citep{marcos}. All of these works operate at a smaller scale (GPT-2 to 11B) and typically on simplistic tasks such as CommonsenseQA, SuperGLUE and multi-digit multiplication. In the VLM domain, \citep{improve_vlm_cot} observe that CoT fine-tuning improves not only CoT-mode but also direct-answer VLM performance, suggesting some degree of internalization. Our work differs from prior implicit CoT methods in several key respects: (i) we operate at 24B--32B scale on a suite of challenging long-document VQA benchmarks, (ii) we achieve internalization through low-strength model merging~\citep{task_arithmetic} without architectural or data modifications, (iii) we demonstrate causal on/off capabilities via a control token, and (iv) we show evidence that reasoning trace design matters in contrast to the finding that incoherent rationales suffice for training-signal enhancement alone~\citep{cot_after_mysteries}.

\section{Method}
\label{sec:method}

\begin{figure}[t]
  \begin{center}
  \includegraphics[width=\linewidth, trim=0 3 0 0, clip]{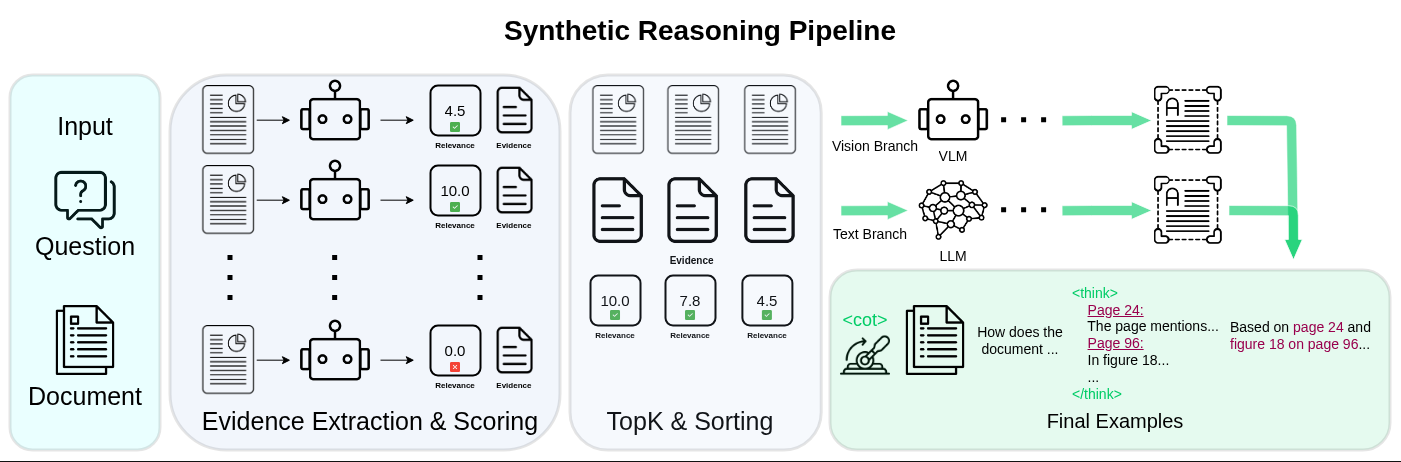}
  \end{center}
  \caption{Our proposed synthetic reasoning pipeline. For a given document and question, we extract evidence and a relevance score from each page using a VLM, then select the top $K$ pages and evidence sections, sort them, then pass them to a strong VLM and LLM respectively to generate an example's final answer. Full examples are constructed with a token controlling the presence of thinking, the document, question, synthetic reasoning trace and the final answer from either branch. }
  \label{fig:pipeline}
\end{figure}

We build on the document corpus, question generation methodology, recursive answer generation pipeline and training infrastructure of \citep{orion}. Questions are generated from three source types within each document---single pages, contiguous page subsets and random page subsets---with prompts selected from an array of question types (math, reasoning, summarization, etc.), targeting information from a single page or multiple pages, providing a range of information-seeking tasks from localized lookup to cross-page synthesis. Given a document of $N$ page images and a question $\mathbf{Q}$ generated from a known subset of source pages, our pipeline produces a structured reasoning trace paired with an answer in two stages: evidence extraction and relevance scoring, and answer generation. Figure~\ref{fig:pipeline} illustrates the complete procedure.

\subsection{Synthetic reasoning trace generation}

\paragraph{Stage 1: evidence extraction and relevance scoring.}
Each page in a document is processed independently by an extractor VLM (Qwen3 VL 32B Instruct), which receives a single page image alongside $\mathbf{Q}$. For each page, the VLM extracts a natural-language evidence snippet with any question-relevant content or visual descriptions and assigns a relevance score on a $[0, 10]$ scale. Pages belonging to the question's source set are explicitly identified to the VLM, guiding it to mark these with a score between $[6.0, 10.0]$. After processing all $N$ pages, we remove page-evidence pairs below a relevance threshold (default $1.0$), then rank them by relevance from most to least relevant and retain up to the top-$K$ pairs (default $K{=}24$). 

\paragraph{Stage 2: answer generation.}
From the ranked evidence set, we generate answers through two parallel branches:
\begin{itemize}
    \item \textbf{Visual branch.} A teacher VLM (Qwen3 VL 235B A22B Instruct) receives the top-ranked page images, typically including the source pages used to generate the question, and $\mathbf{Q}$, generating an answer directly from the visual content without access to the extracted evidence text. 
    \item \textbf{Text branch.} A teacher LLM (Qwen3 235B A22B Instruct\footnote{Released under the Apache 2.0 license.}) \citep{qwen3} receives only the extracted evidence snippets and $\mathbf{Q}$, generating an answer without access to any of the page images.
\end{itemize}
Training examples are drawn equally from both branches. The text branch enforces a dependency between the reasoning trace and the answer, since the answer is generated solely from the evidence that appears in the trace. This ensures the constructed reasoning is causally linked to the answer during training. The visual branch grounds answers in the full visual content, preserving information from charts, figures and tables that text extraction may not fully capture. Both branches benefit from a bounded, relevance-sorted context provided to a strong teacher model, mitigating the impact of context rot \citep{context_rot} and the lost-in-the-middle effect \citep{lost_in_the_middle} respectively. The pipeline can be viewed as RAG, with the extractor VLM acting as a reranker over the document. 

\paragraph{Training example construction.}
Regardless of which branch generated the answer, all training examples follow a unified format. Each page in the document and in the synthetic reasoning is prefixed with its 1-indexed position (\texttt{Page~X:}), making it easy for the model to associate evidence with its source page. The system prompt contains a \texttt{<cot>} control token that gates the reasoning trace: in $95\%$ of examples, \texttt{<cot>} is provided and the reasoning is included; in the remaining $5\%$, the model is trained on the answer alone. This allows us to switch modes at inference-time. For examples with \texttt{<cot>}, the assistant response begins with a \texttt{<think>} block containing the evidence snippets from the top-ranked pages, each prefixed with its page index and ordered from most to least relevant, followed by \texttt{</think>} and the final answer. For the remaining $5\%$ of examples, $\mathbf{Q}$ is followed directly by the final answer. 



\subsection{Trace design}
\label{sec:trace_design}

The structure of the reasoning trace appears to impact its effectiveness. Our initial pipeline (v1) used a naive approach that included every page in document order: pages exceeding a relevance threshold contained their extracted evidence, while remaining pages were labeled \texttt{irrelevant}. This design intended to teach an explicit sequential scan, but it encoded a pathological algorithm. In a typical 100+ page document with only a handful of relevant pages, the vast majority of page extractions are repetitions of \texttt{irrelevant}, and the model observes far more loop-continuation tokens than stopping conditions. We observed that during inference, models trained on v1 traces frequently looped indefinitely, generating page references well past the document length. Given the strength of the v2 pipeline, we hypothesize that even when a low merge strength removes explicit looping, the model internalizes this failure mode and performs poorly as a result.

The v2 redesign reframes the reasoning as a bounded retrieval procedure: only the top-$K$ relevant pages appear, sorted by relevance, with no irrelevant markers. This eliminates the looping failure mode and teaches the model a finite evidence aggregation algorithm---rank pages by relevance, attend to the most informative content, and stop after a bounded number of steps. We show examples of the v1 and v2 trace designs in Figures~\ref{fig:v1_trace_example} and \ref{fig:v2_trace_example}. Additional v2 changes include identifying question source pages to the extractor/scorer, using minimal answer prompting for cleaner distillation and prepending page indices to images to align page references in traces with the visual context. We compare v1 and v2 performance in \S\ref{sec:trace_ablation} but do not isolate each change individually due to resource constraints.

\subsection{Supervision variants and internalization}
\label{sec:three_evidence}

\paragraph{Internalization via model merging.}
All final models are produced by model merging~\citep{task_arithmetic}: $\theta_{\text{merged}} = \theta_{\text{base}} + \alpha \cdot (\theta_{\text{SFT}} - \theta_{\text{base}})$, where $\theta_{\text{base}}$ is the instruction-tuned base model, sometimes merged with the CPT variants from \citep{orion}, and $\alpha$ controls the merge strength. At higher merge strength ($\alpha {=} 0.5$), the model generates explicit reasoning in the majority of responses with substantially longer outputs, achieving comparable scores on MMLBD and MMLBD-C while degrading significantly overall. At low merge strength ($\alpha {=} 0.25$), we find that the model does not generate explicit thinking tokens and mean output length is comparable to the no-think model, yet it retains the full performance benefits of reasoning trace training. 

\paragraph{Control token training.}
We train and evaluate under four settings to disentangle the contribution of the reasoning traces from confounding factors:
\begin{itemize}[leftmargin=1.2em]
    \item \textbf{think}: $95\%$ of examples include the control token \texttt{<cot>} and the full \texttt{<think>} evidence section with the remaining $5\%$ of examples excluding both.
    \item \textbf{no-think}: trained on identical final answers with no control token and the \texttt{<think>} content removed.
    \item \textbf{CoT-on}: evaluation performed \emph{with} \texttt{<cot>} and an expanded generation budget. 
    \item \textbf{CoT-off}: evaluation performed \emph{without} the \texttt{<cot>} token and without an expanded generation budget.
\end{itemize}
Using these and the model merge strength results, we gather three pieces of evidence to argue that the synthetic traces are internalized or compressed by the model merging and the model executes this compressed capability during inference: (i) CoT-on + think substantially outperforms CoT-off + no-think and CoT-on + no-think does not explain the improved performance, showing that the synthetic reasoning is responsible for the performance gains; (ii) CoT-off + think performance degrades, indicating the learned capability is executed during inference; and (iii) interpolating the model along the SFT vector with higher $\alpha$ shifts the model from non-visible to visible reasoning, suggesting the internalized capability derives from the synthetic reasoning trace. 






\section{Experimental setup}

We adopt the data, training procedure and evaluation protocol of \citep{orion} and summarize the key components below.

\paragraph{Evaluation.}
\label{pa:eval}
We evaluate on a suite of long-context benchmarks and report two aggregate metrics:
\begin{description}[nosep,leftmargin=1em,labelindent=0em]
    \item[Visual-LC Avg (VA):] averaged across MMLBD~\citep{mmlbd}, MMLBD-C~\citep{orion}, MMLongBench~\citep{mmlb}, DUDE~\citep{dude} and SlideVQA~\citep{slidevqa}.
    \item[LC Avg (LCA):] VA benchmarks plus HELMET~\citep{helmet} and LongBench v2~\citep{longbench_v2}.
\end{description}
Scores are normalized by the per-benchmark maximum (typically Qwen3 VL 235B A22B) before averaging to ensure balanced comparison across benchmarks with different score distributions. VA and MMLBD/MMLBD-C are our primary metrics, ensuring we avoid overfitting to any single benchmark while focusing on long-document VQA. Across 3 runs, both aggregates are stable: VA has $\sigma = 0.33$ and LCA has $\sigma = 0.24$. See Appendix~\ref{sec:evaluation_benchmarks} for benchmark details and score variance. 
The expanded generation budget of CoT-on is 48K tokens, though empirically this is not exercised, see Figure~\ref{fig:output_length_distributions}.

\paragraph{Data.}
\label{pa:data}
The document corpus consists of a dataset of 250K PDFs (16M pages), augmented with the PDFA English split~\citep{pdfa} (2M PDFs, 18M pages) \citep{orion}. Documents in the first corpus average ${\sim}34$ pages, while the second has a mean of ${\sim}8.6$ pages, with a long tail of documents capped at 336 pages. Synthetic question--answer pairs are generated from subsets of these documents including singular pages, contiguous page subsets and random page subsets. 

We generate 50K synthetic reasoning examples, using Qwen3 VL 32B Instruct as the extractor and Qwen3 (VL) 235B A22B for final answers. We also include 10K examples from each of the following external SFT dataset mixes: Luth \citep{luthsft}, Smoltalk2 \citep{smoltalk2}. See Appendix~\ref{sec:external_sft_composition} for details on the precise external SFT data mix. The dataset averages 57.9 pages per example, with a median of 16.0 pages. For the Qwen3 VL 235B A22B Thinking reasoning trace baseline, we build an equivalent comparison dataset by providing the entire document and using the full response for training.

\paragraph{Training details.}
\label{pa:training_details}
For all training, we train with supervised fine-tuning (SFT) using AdamW \citep{adamw} and the WSD schedule \citep{wsd}. We use sequence packing, avoid truncating sequences, and normalize loss by the total number of assistant tokens. We dynamically scale document resolution when it will not fit entirely within the context, varying the maximum side resolution from $728$ to $1400$. Optimizer hyperparameters are summarized in Table~\ref{tab:training_hyperparams}.

For efficiency, we train in two stages, a short stage with examples of up to 104 pages and a long stage with examples of up to 336 pages. For Mistral, stage~1 forms packed sequences of $128\text{K}$ tokens and stage~2 is $336\text{K}$ tokens. For Qwen3~VL, stage~1 is $128\text{K}$ tokens and stage~2 is $256\text{K}$ tokens. We use ring attention for context parallelism (CP) \citep{ring_attn,ring_flash_attn}. We do not modify the RoPE $\theta$ for either model. Parallelism configurations are shown in Table~\ref{tab:parallelism_config}.

We merge the CPT vectors from \citep{orion} with the respective instruct model using a merge strength of 0.25 for Qwen3 VL and 0.5 for Mistral.

\begin{table}[h]
\begin{center}
\resizebox{\textwidth}{!}{%
\begin{tabular}{lccccccc}
\toprule
\multicolumn{1}{c}{\bf Schedule} & \multicolumn{1}{c}{\bf Max LR} & \multicolumn{1}{c}{\bf Warmup/Decay} & \multicolumn{1}{c}{\bf $\beta_1$} & \multicolumn{1}{c}{\bf $\beta_2$} & \multicolumn{1}{c}{\bf $\epsilon$} & \multicolumn{1}{c}{\bf Weight Decay} & \multicolumn{1}{c}{\bf Grad Clip} \\
\midrule
WSD \citep{wsd} & $5e\text{-}6$ & 10\% samples & 0.9 & 0.999 & $10^{-9}$ & 0.1 & 1.0 \\
\bottomrule
\end{tabular}
}%
\end{center}
\caption{Optimizer hyperparameters.}\label{tab:training_hyperparams}
\end{table}

\begin{table}[h]
\begin{center}
\resizebox{\textwidth}{!}{%
\begin{tabular}{lccccccc}
\toprule
\multicolumn{1}{l}{\bf Stage} & \multicolumn{1}{c}{\bf Hardware} & \multicolumn{1}{c}{\bf CP} & \multicolumn{1}{c}{\bf DP [shard, replicate]} & \multicolumn{1}{c}{\bf Batch Size} & \multicolumn{1}{c}{\bf Tokens per Batch (Qwen/Mistral)} & \multicolumn{1}{c}{\bf Hours (Qwen/Mistral)} \\
\midrule
Stage 1 & H100 & 32 & [32, 2] & 2 & 256K/256K & 23/16 \\
Stage 2 & H100 & 64 & [64, 2] & 2 & 512K/672K & 55/34 \\
\bottomrule
\end{tabular}
}%
\end{center}
\caption{Parallelism and hardware configuration for each stage. CP = context parallelism degree, DP = data parallelism. From this, we estimate our main training experiments require ~40K H100 hours, while additional compute for evaluation, dataset generation and additional experiments bring the project total to $\sim$100K H100 hours. }\label{tab:parallelism_config}
\end{table}

\section{Experiments}
\label{sec:experiments}

We organize experiments around our contributions: state-of-the-art performance via synthetic reasoning (\S\ref{sec:main_results}), internalization of reasoning with minimal token overhead and causal evidence for the inference-time capability via control token ablations (\S\ref{sec:internalization}), and the impact of trace design (\S\ref{sec:trace_ablation}).

\subsection{Main results}
\label{sec:main_results}

\begin{table}[t]
\centering
\begin{tabular}{lcc}
\toprule
\bf Model & \bf Acc & \bf Params \\
\midrule
\bf Synthetic Reasoning Qwen (Ours) & \bf 58.3 & 32B \\
\midrule
Qwen3 VL 235B A22B Instruct & 57.0 & 235B (22B) \\
Qwen3 VL 235B A22B Thinking & 56.2 & 235B (22B) \\
TeleMM-2.0 & 56.1 & -- \\
Qwen3 VL 32B Instruct & 55.4 & 32B \\
GLM 4.6V & 54.9 & 106B (12B) \\
GPT-4o & 46.3 & -- \\
\bottomrule
\end{tabular}
\vspace{0.5em}
\caption{Official MMLongBenchDoc leaderboard (accuracy). Our 32B model surpasses models with 7$\times$ more parameters.}\label{tab:leaderboard_and_output}
\end{table}

Table~\ref{tab:leaderboard_and_output} presents official MMLongBenchDoc leaderboard results. Our synthetic reasoning model achieves state-of-the-art results with 58.3 accuracy, surpassing Qwen3 VL 235B A22B Instruct (57.0) with 7$\times$ fewer parameters. We note the competitive nature of the benchmark: previous to Qwen3 VL in September 2025, no model surpassed GPT-4o. Strong models fell well short of the human-expert baseline at 65.8 and while the gap has narrowed significantly, recent models have saturated around Qwen3 VL 235B's score. We improve Qwen3 VL 32B's score by +2.9 points, which is made substantial by this saturation. Additionally, we observe that the existing reasoning capabilities of Qwen3 VL models do not improve document understanding: the Thinking variant scores slightly below its Instruct counterpart, motivating our novel synthetic reasoning design.

\begin{table}[t]
  \begin{center}
  \resizebox{\textwidth}{!}{%
  \begin{tabular}{lccccccccccc}
  \toprule
  \multicolumn{1}{l}{\bf Model} & \multicolumn{1}{c}{\bf VA} & \multicolumn{1}{c}{\bf LCA} & \multicolumn{1}{c}{\bf MMLBD} & \multicolumn{1}{c}{\bf MMLBD-C} & \multicolumn{1}{c}{\bf MMLB 128K} & \multicolumn{1}{c}{\bf MMLB 32K} & \multicolumn{1}{c}{\bf SlideVQA} & \multicolumn{1}{c}{\bf Helmet} & \multicolumn{1}{c}{\bf LongBench v2} & \multicolumn{1}{c}{\bf DUDE} & \multicolumn{1}{c}{\bf TableVQA} \\
  \midrule
  \multicolumn{12}{l}{\bf Qwen3 VL} \\
  ~~235B A22B Instruct & \textbf{98.4} & \textbf{98.5} & 54.8 & 56.2 & \textbf{78.6} & \textbf{82.4} & \textbf{84.5} & 67.6 & \textbf{50.0} & 59.1 & 78.8 \\
  ~~\textbf{Synthetic Reasoning (Ours)} & 95.0 \textcolor{teal}{(+1.3)} & 94.4 \textcolor{teal}{(+2.3)} & \textbf{55.8} \textcolor{teal}{(+4.0)} & \textbf{58.2} \textcolor{teal}{(+4.4)} & 75.7 \textcolor{teal}{(+5.3)} & 78.6 \textcolor{red}{(-0.3)} & 75.4 \textcolor{red}{(-1.8)} & \textbf{68.5} \textcolor{teal}{(+5.5)} & 43.0 \textcolor{teal}{(+1.0)} & 55.1 \textcolor{red}{(-6.7)} & 81.6 \textcolor{teal}{(+4.0)} \\
  ~~LongPO \citep{orion} & 94.0 \textcolor{teal}{(+0.3)} & 92.4 \textcolor{teal}{(+0.3)} & 53.6 \textcolor{teal}{(+1.8)} & 56.4 \textcolor{teal}{(+2.6)} & 75.6 \textcolor{teal}{(+5.2)} & 78.4 \textcolor{red}{(-0.5)} & 75.5 \textcolor{red}{(-1.7)} & 62.9 \textcolor{red}{(-0.1)} & 42.0 & 56.0 \textcolor{red}{(-5.8)} & 81.7 \textcolor{teal}{(+4.1)} \\
  ~~32B Instruct & 93.7 & 92.1 & 51.8 & 53.8 & 70.4 & 78.9 & 77.2 & 63.0 & 42.0 & \textbf{61.8} & 77.6 \\
  ~~Plain Distillation \citep{orion} & 92.0 \textcolor{red}{(-1.7)} & 91.8 \textcolor{red}{(-0.3)} & 54.9 \textcolor{teal}{(+3.1)} & 57.3 \textcolor{teal}{(+3.5)} & 73.8 \textcolor{teal}{(+3.4)} & 77.0 \textcolor{red}{(-1.9)} & 66.8 \textcolor{red}{(-10.4)} & 65.7 \textcolor{teal}{(+2.7)} & 44.0 \textcolor{teal}{(+2.0)} & 54.8 \textcolor{red}{(-7.0)} & \textbf{82.1} \textcolor{teal}{(+4.5)} \\
  ~~No-think & 91.7 \textcolor{red}{(-2.0)} & 91.9 \textcolor{red}{(-0.2)} & 52.2 \textcolor{teal}{(+0.4)} & 54.5 \textcolor{teal}{(+0.7)} & 72.0 \textcolor{teal}{(+1.6)} & 77.7 \textcolor{red}{(-1.2)} & 74.2 \textcolor{red}{(-3.0)} & 65.9 \textcolor{teal}{(+2.9)} & 45.0 \textcolor{teal}{(+3.0)} & 55.3 \textcolor{red}{(-6.5)} & 81.7 \textcolor{teal}{(+4.1)} \\
  \midrule
  \multicolumn{12}{l}{\bf Mistral} \\
  ~~\textbf{Synthetic Reasoning (Ours)} & 87.8 \textcolor{teal}{(+7.6)} & 87.9 \textcolor{teal}{(+11.3)} & 47.5 \textcolor{teal}{(+7.6)} & 49.3 \textcolor{teal}{(+7.9)} & 75.4 \textcolor{teal}{(+9.0)} & 75.0 \textcolor{teal}{(+2.1)} & 69.7 \textcolor{teal}{(+1.9)} & 62.6 \textcolor{teal}{(+25.6)} & 43.0 \textcolor{teal}{(+4.0)} & 54.1 \textcolor{teal}{(+1.3)} & 77.6 \textcolor{teal}{(+23.6)} \\
  ~~Plain Distillation \citep{orion} & 84.4 \textcolor{teal}{(+4.2)} & 82.4 \textcolor{teal}{(+5.8)} & 46.6 \textcolor{teal}{(+6.7)} & 47.4 \textcolor{teal}{(+6.0)} & 65.7 \textcolor{red}{(-0.7)} & 71.3 \textcolor{red}{(-1.6)} & 71.2 \textcolor{teal}{(+3.4)} & 53.1 \textcolor{teal}{(+16.1)} & 38.0 \textcolor{red}{(-1.0)} & 54.0 \textcolor{teal}{(+1.2)} & 76.6 \textcolor{teal}{(+22.6)} \\
  ~~Qwen Thinking Traces & 82.6 \textcolor{teal}{(+2.4)} & 82.7 \textcolor{teal}{(+6.1)} & 45.0 \textcolor{teal}{(+5.1)} & 45.5 \textcolor{teal}{(+4.1)} & 60.6 \textcolor{red}{(-5.8)} & 74.2 \textcolor{teal}{(+1.3)} & 69.4 \textcolor{teal}{(+1.6)} & 64.1 \textcolor{teal}{(+27.1)} & 37.0 \textcolor{red}{(-2.0)} & 54.0 \textcolor{teal}{(+1.2)} & 76.7 \textcolor{teal}{(+22.7)} \\
  ~~3.1 Small 24B & 80.2 & 76.6 & 39.9 & 41.4 & 66.4 & 72.9 & 67.8 & 37.0 & 39.0 & 52.8 & 54.0 \\
  ~~No-think & 77.9 \textcolor{red}{(-2.3)} & 77.7 \textcolor{teal}{(+1.1)} & 43.0 \textcolor{teal}{(+3.1)} & 44.3 \textcolor{teal}{(+2.9)} & 49.2 \textcolor{red}{(-17.2)} & 72.2 \textcolor{red}{(-0.7)} & 68.4 \textcolor{teal}{(+0.6)} & 55.8 \textcolor{teal}{(+18.8)} & 37.0 \textcolor{red}{(-2.0)} & 51.4 \textcolor{red}{(-1.4)} & 75.5 \textcolor{teal}{(+21.5)} \\
  \bottomrule
  \end{tabular}
  }%
  \end{center}
  \caption{Comparison of the synthetic reasoning models to the base models and to baseline models trained on thinking traces from Qwen3 VL 235B A22B Thinking on our benchmark suite, with deltas shown relative to the base model. We include the best performing baselines from \citep{orion} for additional perspective. Plain Distillation uses answers generated with Qwen3 VL 235B A22B Instruct with the whole document as context.}\label{tab:main_comparisons}
\end{table}

Table~\ref{tab:main_comparisons} presents results across our full benchmark suite. Synthetic reasoning achieves the highest VA and LCA for both model families. For Qwen on the primary long-document benchmarks, synthetic reasoning achieves 55.8 MMLBD and 58.2 MMLBD-C, surpassing LongPO \citep{orion}, the no-think baseline and Qwen3 VL 235B A22B Instruct. For Mistral, the gains are more impressive: +7.9 MMLBD-C over the base model and synthetic reasoning nearly doubles the gains over the base model relative to plain distillation SFT in VA and LCA. 
In both cases, we see the synthetic reasoning model is more effective than the non-reasoning version.

\subsection{Internalization and control token analysis}
\label{sec:internalization}


We have verified the think model is more effective than no-think. If our synthetic reasoning method is responsible for performance gains, then we also expect the expanded generation budget and the \texttt{<cot>} token, which may simply encourage the model to reason harder regardless of training, not to recover the performance of CoT-on + think model. As shown in Table~\ref{tab:control_token_results}, the only consistent effect across models is an increase in HELMET scores. For both models, CoT-on + no-think does not approach CoT-on + think performance (evidence (i) from \S\ref{sec:three_evidence}).

Interested in this finding, we investigate the effect of removing the \texttt{<cot>} control token from the think model during evaluation. As shown in Table~\ref{tab:control_token_results}, compared to CoT-on + think, the CoT-off + think setting degrades significantly for both models. Thus, during inference we can control the answer quality, with the model performing better when in 'reasoning mode' (evidence (ii) from \S\ref{sec:three_evidence}). 

\begin{figure}[t]
  \centering
  \begin{minipage}[c]{0.59\linewidth}
    \centering
    \includegraphics[width=\linewidth]{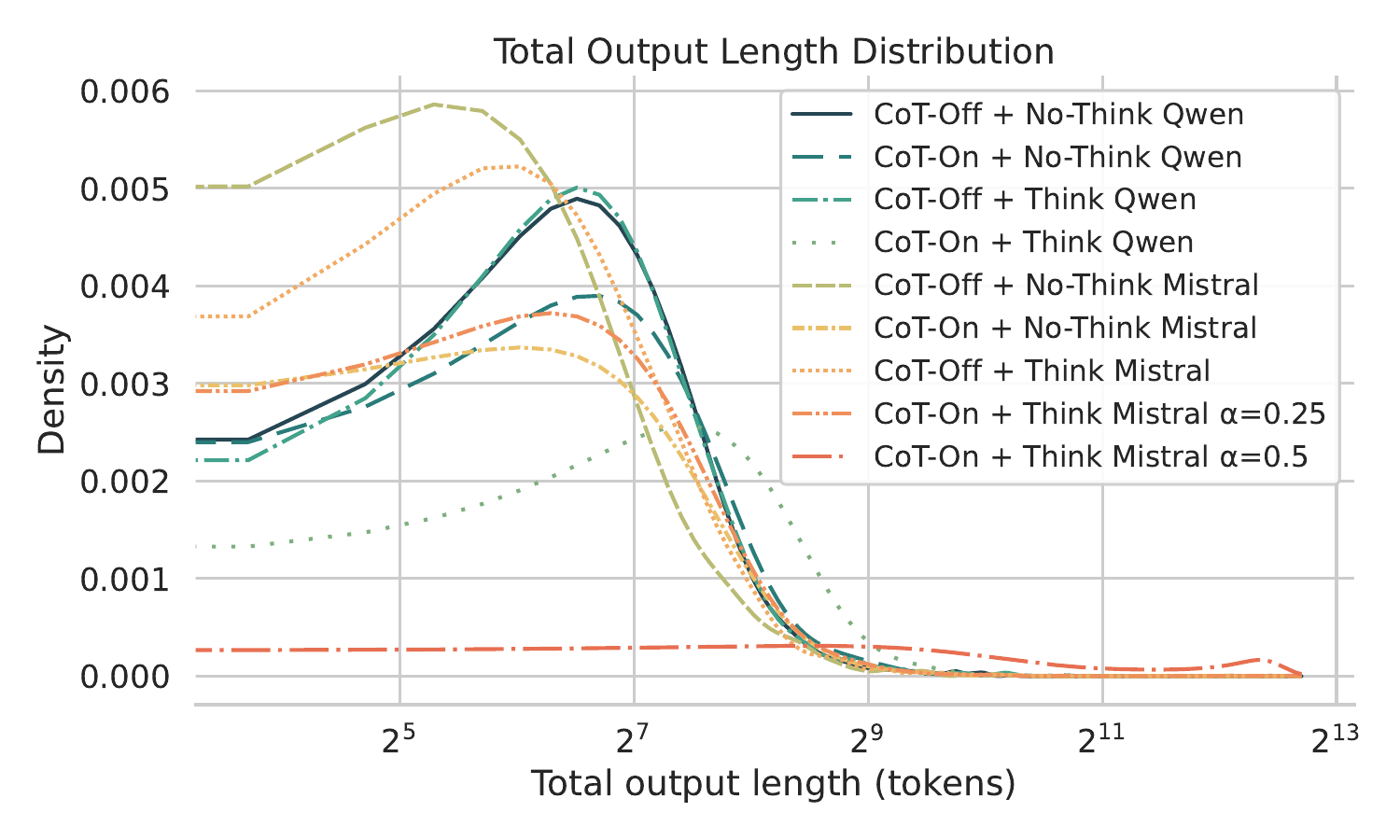}
    \captionof{figure}{Output length distributions for Qwen and Mistral under different evaluation + train settings on MMLBD. Generally we see CoT-on produces longer-tailed distributions, but minimally compared to the $\alpha = 0.5$ model which explicitly reasons. Additionally, the control token determines output length distribution more than think vs no-think training.}
    \label{fig:output_length_distributions}
  \end{minipage}
  \hfill
  \begin{minipage}[c]{0.38\linewidth}
    \centering
    \scriptsize
    \begin{tabular}{@{}p{0.72\linewidth}r@{}}
    \toprule
    \bf Setting & \bf Mean Tokens \\
    \midrule
    \multicolumn{2}{l}{\bf Qwen3 VL} \\
    ~~Base model & 124 \\
    ~~no-think & 141 \\
    ~~Syn. Reasoning 0.25 & 250 \\
    \midrule
    \multicolumn{2}{l}{\bf Mistral} \\
    ~~Base model & 61 \\
    ~~no-think & 102 \\
    ~~Syn. Reasoning 0.25 & 132 \\
    ~~Syn. Reasoning 0.5 & 1637 \\
    \bottomrule
    \end{tabular}
    \captionof{table}{Mean output tokens on MMLBD. Explicit reasoning ($\alpha = 0.5$) averages 12.4$\times$ more output tokens than implicit ($\alpha = 0.25$).}
    \label{tab:output_length_stats}
  \end{minipage}
\end{figure}

Finally, from Table~\ref{tab:output_length_stats} and Figure~\ref{fig:output_length_distributions}, we see our final models at merge strength $\alpha = 0.25$ do not exhibit explicit thinking, i.e. the \texttt{<think>} block is absent in generated responses and, additionally, the reasoning has not simply migrated outside this block. As we interpolate further along the SFT vector, with $\alpha = 0.5$, the model shifts from minimal token overhead, internalized reasoning, to externalized, visible reasoning (evidence (iii) from \S\ref{sec:three_evidence}). We interpret that the low merge strength model internalizes the synthetic reasoning process, which is evoked visibly as the strength of the SFT is increased.

Collectively, we argue the model has internalized the synthetic reasoning or a computation learned from it, which it can selectively execute during inference, rather than that the model is merely better trained because of the synthetic reasoning trace. More specifically, CoT faithfulness literature shows that CoTs can, depending on the setup, be unimportant rationalizations~\citep{reasoning_theater} or partially 'load-bearing'~\citep{cot_program_variables} in the sense that modifying load-bearing tokens changes the final answer. In our case, we know the synthetic reasoning trace as a whole is load-bearing, but it is likely that explicitly executing the CoT is not necessary for the performance gains, i.e. the model learns an algorithm from it that is compressible. Our generally applicable finding is that merge strength $\alpha$ can be used to achieve this compression, yielding a method for internalizing expensive reasoning, such as our verbose synthetic reasoning algorithm, while retaining performance. We leave the investigation of this method in other reasoning settings for future work.

\paragraph{Thinking traces vs. synthetic reasoning.}
Synthetic reasoning also outperforms training Mistral on thinking traces from Qwen3 VL 235B A22B Thinking by +3.8 MMLBD-C (Table~\ref{tab:main_comparisons}). Interestingly, Table~\ref{tab:control_token_results} shows that the thinking-trace model exhibits minimal difference between CoT-on and CoT-off evaluation; we hypothesize the thinking traces do not substantially affect performance. The model may have internalized the reasoning in the same fashion as our synthetic traces, but with little measurable impact. This is in line with our observation of the similar performance of Qwen3 VL 235B A22B Thinking and Instruct on MMLongBenchDoc. 

\begin{table}[t]
  \begin{center}
  \resizebox{\textwidth}{!}{%
  \begin{tabular}{lccccccccccc}
  \toprule
  \multicolumn{1}{l}{\bf Eval + Train Setting} & \multicolumn{1}{c}{\bf VA} & \multicolumn{1}{c}{\bf LCA} & \multicolumn{1}{c}{\bf MMLBD} & \multicolumn{1}{c}{\bf MMLBD-C} & \multicolumn{1}{c}{\bf MMLB 128K} & \multicolumn{1}{c}{\bf MMLB 32K} & \multicolumn{1}{c}{\bf SlideVQA} & \multicolumn{1}{c}{\bf Helmet} & \multicolumn{1}{c}{\bf LongBench v2} & \multicolumn{1}{c}{\bf DUDE} & \multicolumn{1}{c}{\bf TableVQA} \\
  \midrule
  \multicolumn{12}{l}{\bf Qwen3 VL} \\
  ~~CoT-on + think & \textbf{95.0} \textcolor{teal}{(+3.3)} & \textbf{94.4} \textcolor{teal}{(+2.5)} & \textbf{55.8} \textcolor{teal}{(+3.6)} & \textbf{58.2} \textcolor{teal}{(+3.7)} & \textbf{75.7} \textcolor{teal}{(+3.7)} & \textbf{78.6} \textcolor{teal}{(+0.9)} & 75.4 \textcolor{teal}{(+1.2)} & 68.5 \textcolor{teal}{(+2.6)} & 43.0 \textcolor{red}{(-2.0)} & 55.1 \textcolor{red}{(-0.2)} & \textbf{81.6} \textcolor{red}{(-0.1)} \\
  ~~CoT-off + think & 93.2 \textcolor{teal}{(+1.5)} & 92.0 \textcolor{teal}{(+0.1)} & 52.4 \textcolor{teal}{(+0.2)} & 54.9 \textcolor{teal}{(+0.4)} & 74.8 \textcolor{teal}{(+2.8)} & 78.2 \textcolor{teal}{(+0.5)} & \textbf{75.9} \textcolor{teal}{(+1.7)} & 64.3 \textcolor{red}{(-1.6)} & 42.0 \textcolor{red}{(-3.0)} & \textbf{56.2} \textcolor{teal}{(+0.9)} & 79.8 \textcolor{red}{(-1.9)} \\
  ~~CoT-off + no-think & 91.7 & 91.9 & 52.2 & 54.5 & 72.0 & 77.7 & 74.2 & 65.9 & \textbf{45.0} & 55.3 & 81.7 \\
  ~~CoT-on + no-think & 91.0 \textcolor{red}{(-0.7)} & 91.2 \textcolor{red}{(-0.7)} & 51.4 \textcolor{red}{(-0.8)} & 53.2 \textcolor{red}{(-1.3)} & 72.3 \textcolor{teal}{(+0.3)} & 77.3 \textcolor{red}{(-0.4)} & 74.0 \textcolor{red}{(-0.2)} & \textbf{69.3} \textcolor{teal}{(+3.4)} & 42.0 \textcolor{red}{(-3.0)} & 55.0 \textcolor{red}{(-0.3)} & 80.0 \textcolor{red}{(-1.7)} \\
  \midrule
  \multicolumn{12}{l}{\bf Mistral} \\
  ~~CoT-on + think & 87.8 \textcolor{teal}{(+9.9)} & 87.9 \textcolor{teal}{(+10.2)} & 47.5 \textcolor{teal}{(+4.5)} & 49.3 \textcolor{teal}{(+5.0)} & 75.4 \textcolor{teal}{(+26.2)} & 75.0 \textcolor{teal}{(+2.8)} & 69.7 \textcolor{teal}{(+1.3)} & 62.6 \textcolor{teal}{(+6.8)} & 43.0 \textcolor{teal}{(+6.0)} & 54.1 \textcolor{teal}{(+2.7)} & 77.6 \textcolor{teal}{(+2.1)} \\
  ~~CoT-on + no-think & 83.8 \textcolor{teal}{(+5.9)} & 83.5 \textcolor{teal}{(+5.8)} & 42.4 \textcolor{red}{(-0.6)} & 44.8 \textcolor{teal}{(+0.5)} & \textbf{75.7} \textcolor{teal}{(+26.5)} & 74.8 \textcolor{teal}{(+2.6)} & 67.3 \textcolor{red}{(-1.1)} & 61.7 \textcolor{teal}{(+5.9)} & 38.0 \textcolor{teal}{(+1.0)} & 51.4 & 76.3 \textcolor{teal}{(+0.8)} \\
  ~~CoT-off + think & 82.0 \textcolor{teal}{(+4.1)} & 82.9 \textcolor{teal}{(+5.2)} & 45.4 \textcolor{teal}{(+2.4)} & 49.2 \textcolor{teal}{(+4.9)} & 56.5 \textcolor{teal}{(+7.3)} & 70.4 \textcolor{red}{(-1.8)} & 68.8 \textcolor{teal}{(+0.4)} & 56.3 \textcolor{teal}{(+0.5)} & \textbf{45.0} \textcolor{teal}{(+8.0)} & 53.8 \textcolor{teal}{(+2.4)} & 77.6 \textcolor{teal}{(+2.1)} \\
  ~~CoT-off + no-think & 77.9 & 77.7 & 43.0 & 44.3 & 49.2 & 72.2 & 68.4 & 55.8 & 37.0 & 51.4 & 75.5 \\
  \midrule
  \multicolumn{12}{l}{\bf Mistral (Qwen Thinking Traces)} \\
  ~~CoT-on + think & 82.6 \textcolor{teal}{(+0.3)} & 82.7 \textcolor{teal}{(+1.7)} & 45.0 \textcolor{teal}{(+0.7)} & 45.5 \textcolor{teal}{(+0.2)} & 60.6 \textcolor{red}{(-2.1)} & 74.2 \textcolor{teal}{(+3.3)} & 69.4 \textcolor{red}{(-0.9)} & 64.1 \textcolor{teal}{(+4.3)} & 37.0 \textcolor{teal}{(+3.0)} & 54.0 \textcolor{teal}{(+0.2)} & 76.7 \textcolor{red}{(-0.6)} \\
  ~~CoT-off + think & 82.3 & 81.0 & 44.3 & 45.3 & 62.7 & 70.9 & 70.3 & 59.8 & 34.0 & 53.8 & 77.3 \\
  \bottomrule
  \end{tabular}
  }%
  \end{center}
  \caption{Control token ablations on Qwen and Mistral, deltas relative to CoT-off + no-think models. These results affirm our earlier expectations, showing that reasoning improves performance (+3.3 and +9.9 VA for Qwen and Mistral respectively) and the capability learned from reasoning training is active and controllable at inference-time. CoT-on + no-think does not recover CoT-on + think performance. }\label{tab:control_token_results}
\end{table}

\begin{table}[t]
  \begin{center}
  \resizebox{\textwidth}{!}{%
  \begin{tabular}{lccccccccccc}
  \toprule
  \multicolumn{1}{l}{\bf Merge Strength $\alpha$} & \multicolumn{1}{c}{\bf VA} & \multicolumn{1}{c}{\bf LCA} & \multicolumn{1}{c}{\bf MMLBD} & \multicolumn{1}{c}{\bf MMLBD-C} & \multicolumn{1}{c}{\bf MMLB 128K} & \multicolumn{1}{c}{\bf MMLB 32K} & \multicolumn{1}{c}{\bf SlideVQA} & \multicolumn{1}{c}{\bf Helmet} & \multicolumn{1}{c}{\bf LongBench v2} & \multicolumn{1}{c}{\bf DUDE} & \multicolumn{1}{c}{\bf TableVQA} \\
  \midrule
  0.25 & 87.8 \textcolor{teal}{(+4.7)} & 87.9 \textcolor{teal}{(+9.7)} & 47.5 \textcolor{red}{(-0.1)} & 49.3 \textcolor{teal}{(+1.1)} & 75.4 \textcolor{teal}{(+6.9)} & 75.0 \textcolor{teal}{(+8.8)} & 69.7 \textcolor{teal}{(+1.4)} & 62.6 \textcolor{teal}{(+24.5)} & 43.0 \textcolor{teal}{(+7.0)} & 54.1 \textcolor{teal}{(+3.4)} & 77.6 \textcolor{red}{(-2.1)} \\
  0.5 & 83.1 & 78.2 & 47.6 & 48.2 & 68.5 & 66.2 & 68.3 & 38.1 & 36.0 & 50.7 & 79.7 \\
  \bottomrule
  \end{tabular}
  }%
  \end{center}
  \caption{A comparison of the performance of synthetic reasoning with different merge strengths with Mistral. The performance while explicitly reasoning is comparable to the $\alpha = 0.25$ model in the targeted area, MMLBD and MMLBD-C, despite high degradation in other areas. This indicates the internalized reasoning is highly effective with minimal overhead (Table~\ref{tab:output_length_stats}). }\label{tab:merge_strength_results}
\end{table}

\subsection{Trace iteration}
\label{sec:trace_ablation}

\begin{table}[t]
  \begin{center}
  \resizebox{\textwidth}{!}{%
  \begin{tabular}{lccccccccccc}
  \toprule
  \multicolumn{1}{l}{\bf Checkpoint} & \multicolumn{1}{c}{\bf VA} & \multicolumn{1}{c}{\bf LCA} & \multicolumn{1}{c}{\bf MMLBD} & \multicolumn{1}{c}{\bf MMLBD-C} & \multicolumn{1}{c}{\bf MMLB 128K} & \multicolumn{1}{c}{\bf MMLB 32K} & \multicolumn{1}{c}{\bf SlideVQA} & \multicolumn{1}{c}{\bf Helmet} & \multicolumn{1}{c}{\bf LongBench v2} & \multicolumn{1}{c}{\bf DUDE} & \multicolumn{1}{c}{\bf TableVQA} \\
  \midrule
  Synthetic Reasoning V2 & 87.8 \textcolor{teal}{(+4.3)} & 87.9 \textcolor{teal}{(+2.8)} & 47.5 \textcolor{teal}{(+3.1)} & 49.3 \textcolor{teal}{(+4.2)} & 75.4 \textcolor{teal}{(+6.2)} & 75.0 \textcolor{teal}{(+1.9)} & 69.7 \textcolor{teal}{(+0.3)} & 62.6 \textcolor{red}{(-2.1)} & 43.0 & 54.1 \textcolor{teal}{(+1.4)} & 77.6 \textcolor{teal}{(+0.2)} \\
  Synthetic Reasoning V1 & 83.5 & 85.1 & 44.4 & 45.1 & 69.2 & 73.1 & 69.4 & 64.7 & 43.0 & 52.7 & 77.4 \\
  \bottomrule
  \end{tabular}
  }%
  \end{center}
  \caption{Synthetic reasoning v1 vs v2 on Mistral. The v2 trace redesign (\S\ref{sec:trace_design}) yields substantial improvements across all primary metrics.}\label{tab:v1_vs_v2}
\end{table}

The v1-to-v2 trace redesign (\S\ref{sec:trace_design}) accounts for substantial performance gains on Mistral in VA, MMLBD and MMLBD-C (Table~\ref{tab:v1_vs_v2}). The v1 traces encoded a sequential scan over all pages and the model frequently looped indefinitely during inference. The v2 redesign eliminates the looping failure mode and challenges the model to explicitly consider page relevance and content. 

We highlight that the comparison to no-think baselines, to frontier-model thinking traces and of our bounded, relevance-ordered design over our simple for-loop design (\S\ref{sec:trace_ablation}) contribute three axes of evidence to the content of our synthetic traces: our reasoning is not only \textit{effective}, in comparison to no reasoning, but significantly more effective than Qwen's reasoning and the alternative design.

\section{Limitations}

The v1-to-v2 trace redesign bundled multiple changes, including question source page identification, bounded top-$K$ selection, relevance ordering, page index prefixes and minimal answer prompting, and we do not isolate the contribution of each individual design choice. Further, prompting the extractor with the question source pages may not be viable for real user queries and the impact of this is not measured. Our evaluation relies on a local judge (GLM 4.5V) and subsamples expensive/long-running benchmarks (20 samples per task for HELMET and MMLongBench; see Appendix~\ref{sec:evaluation_benchmarks}), which introduces variance relative to official scoring. For MMLongBenchDoc, we report official leaderboard results separately. For training, we use SFT alone; whether combining synthetic reasoning traces with reinforcement learning methods such as preference optimization (e.g., LongPO) yields further gains is an open and natural question. Finally, our evidence for internalization is behavioral: we establish controllable on/off behavior via control token ablations and output length analysis, but we do not explain the mechanism by which the model represents the internalized capability.

\section{Conclusion}
\label{sec:conclusion}


We propose a synthetic data pipeline for reasoning in long-document VQA and show that this reasoning can be internalized into a low-overhead, token-controlled capability which achieves SOTA on MMLongBenchDoc (58.3, surpassing a 7$\times$ larger model) when active. To understand this behavior, we investigate various control token ablations and model merge strengths and show that our synthetic reasoning is highly effective, is active during inference even when internalized and a low merge strength model compresses or internalizes the synthetic reasoning process, yielding a generalizable method for compressing expensive reasoning while retaining performance. Furthermore, our synthetic traces outperform training on free-form thinking traces from Qwen3 VL 235B A22B Thinking, indicating that the design of the trace, not merely its presence, determines its effectiveness for long-document understanding. These findings suggest that frontier reasoning and non-reasoning models alike can benefit from structured synthetic reasoning in domains where verifiable rewards are scarce.


\section*{Acknowledgements}
We thank Oskar Hallstr\"om for his suggestions on model merging and valuable assistance with experiment design. We also thank the LightOn team for their support and feedback throughout the project.

This work was granted access to the HPC resources of IDRIS under the allocations \texttt{AS011016449} and \texttt{A0181016214} made by GENCI enabling us to use the Jean Zay supercomputer. We thank the IDRIS support team for their valuable help.

This project also received funding from the BPI Scribe project.

We acknowledge EuroHPC JU for awarding the project ID EHPC-AIF-2025FL01-523 access to MareNostrum5 at BSC, Spain.

\bibliographystyle{plainnat}
\bibliography{references}

\clearpage
\appendix

\section{Evaluation}
\label{sec:evaluation_benchmarks}

We follow the evaluation settings of \citep{orion}, which we recall here. We evaluate on a suite of long-context benchmarks spanning document VQA and long-context text tasks. Specifically, we include MMLBD \citep{mmlbd} (and the corrected variant MMLBD-C \citep{orion}); MMLongBench \citep{mmlb} at 32K and 128K context (document QA, visual RAG, ICL, summarization); SlideVQA Mini \citep{slidevqa}; HELMET \citep{helmet} at 32K and 128K context (recall, RAG, summarization, ICL, reranking); LongBench v2 \citep{longbench_v2}; DUDE Mini \citep{dude}; TableVQA \citep{tablevqa}. In contrast to the default VLM Eval Kit \citep{vlmevalkit} settings, we increase the maximum number of pages from 120 to 336 for MMLBD and MMLBD-C, and we set the maximum resolution to $1024 \times 1024$ to ensure long examples fit in context while preserving fine details. We list the specific metrics used for each benchmark below. Due to the large number of evaluations, we limit expensive benchmarks (HELMET and MMLongBench) to 20 samples per task, and we use a local judge, selecting GLM 4.5V due to its strong performance on MMLBD and LM Arena \citep{lmarena} while being outside the model families we train to avoid self-preference bias \citep{selfpreference}. GLM 4.5V was selected in \citep{orion} before the release of 4.6V and we adopt the same evaluation protocol for fair comparison. The local judge and use of F1 are the main factors driving the score difference between official MMLBD results and the ones reported in our paper. 

Table~\ref{tab:benchmark_metrics} lists the primary metric used for each benchmark in our evaluation suite. We release an html file with the full set of scores for easy exploration.

\begin{table}[t]
  \centering
  \resizebox{\textwidth}{!}{%
  \begin{tabular}{ll}
  \toprule
  \textbf{Benchmark} & \textbf{Metric} \\
  \midrule
  MMLongBenchDoc / MMLBD-C & F1 (overall\_f1) \\
  MMLongBench (32K/128K) & Avg of task-specific metrics$^*$ \\
  SlideVQA Mini & ANLS (Average Normalized Levenshtein Similarity) \\
  HELMET (32K/128K) & Overall Score \\
  LongBench v2 & Overall Accuracy \\
  DUDE Mini & ANLS (Average Normalized Levenshtein Similarity) \\
  TableVQA & Average Accuracy across subtasks \\
  \bottomrule
  \end{tabular}
  }%
  \caption{Primary metrics used for each benchmark. All scores are normalized to 0--100 before averaging. $^*$MMLongBench task-specific metrics: Visual RAG (infoseek, viquae): sub\_em, ICL (cars196, food101, inat2021, sun397): cls\_acc, Summarization (gov-report, multi-lexsum): judge\_f1.}
  \label{tab:benchmark_metrics}
\end{table}

In Table \ref{tab:eval_variance}, we show results for the variance of VA and LCA across 3 runs from \citep{orion}. The aggregate metrics are stable across runs, with $\sigma = 0.33$ for VA and $\sigma = 0.24$ for LCA. However, we note that the variance of MMLongBench is especially high, with $\sigma = 1.66$. This is likely due to limiting the number of examples to 20 per task, with a total of $180$ examples for each context length ($32K$ and $128K$).

\begin{table}[t]
  \begin{center}
  \resizebox{\textwidth}{!}{%
  \begin{tabular}{lccccccccccc}
  \toprule
  \multicolumn{1}{l}{\bf Run} & \multicolumn{1}{c}{\bf VA} & \multicolumn{1}{c}{\bf LCA} & \multicolumn{1}{c}{\bf MMLBD} & \multicolumn{1}{c}{\bf MMLBD-C} & \multicolumn{1}{c}{\bf MMLB 128K} & \multicolumn{1}{c}{\bf MMLB 32K} & \multicolumn{1}{c}{\bf SlideVQA} & \multicolumn{1}{c}{\bf Helmet} & \multicolumn{1}{c}{\bf LongBench v2} & \multicolumn{1}{c}{\bf DUDE} & \multicolumn{1}{c}{\bf TableVQA} \\
  \midrule
  Eval\#3 & 92.3 \textcolor{teal}{(+0.4)} & 91.2 \textcolor{teal}{(+0.4)} & 50.9 \textcolor{red}{(-0.5)} & 54.1 \textcolor{teal}{(+0.1)} & 74.4 \textcolor{teal}{(+2.6)} & 80.1 \textcolor{red}{(-1.4)} & 74.3 \textcolor{red}{(-0.5)} & 62.2 \textcolor{red}{(-1.2)} & 43.0 \textcolor{teal}{(+1.0)} & 55.5 \textcolor{teal}{(+1.4)} & 80.7 \\
  Eval\#1 & 91.9 & 90.8 & 51.4 & 54.0 & 71.8 & 81.5 & 74.8 & 63.4 & 42.0 & 54.1 & 80.7 \\
  Eval\#2 & 91.5 \textcolor{red}{(-0.4)} & 90.5 \textcolor{red}{(-0.3)} & 51.2 \textcolor{red}{(-0.2)} & 53.2 \textcolor{red}{(-0.8)} & 75.8 \textcolor{teal}{(+4.0)} & 76.9 \textcolor{red}{(-4.6)} & 73.2 \textcolor{red}{(-1.6)} & 63.2 \textcolor{red}{(-0.2)} & 42.0 & 55.2 \textcolor{teal}{(+1.1)} & 80.5 \textcolor{red}{(-0.2)} \\
  \bottomrule
  \end{tabular}
  }%
  \end{center}
  \caption{Evaluation variance across 3 runs from \citep{orion}.}\label{tab:eval_variance}
\end{table}

\section{Synthetic reasoning text vs visual branch}
\label{sec:text_vs_visual_branch}

With V1 of synthetic reasoning, we trained two models, one with answers from the text branch and one with answers from the visual branch. We find that the text branch is slightly better than the visual branch (+1.1 VA) and significantly better on MMLongBenchDoc and MMLBD-C (+2.3 and +2.0 MMLBD-C respectively). Given the changes to improve the visual branch in V2 (see \S\ref{sec:trace_design}) and a desire to balance the visual strength of the model, we increased the portion of answers sampled from the text branch in V2 from 40\% to 50\%. We find it interesting that the text branch performs so well, given access to only the extracted evidence, indicating the extracted evidence contains enough information to answer questions effectively.

\begin{table}[t]
  \begin{center}
  \resizebox{\textwidth}{!}{%
  \begin{tabular}{lccccccccccc}
  \toprule
  \multicolumn{1}{l}{\bf Answer Branch} & \multicolumn{1}{c}{\bf VA} & \multicolumn{1}{c}{\bf LCA} & \multicolumn{1}{c}{\bf MMLBD} & \multicolumn{1}{c}{\bf MMLBD-C} & \multicolumn{1}{c}{\bf MMLB 128K} & \multicolumn{1}{c}{\bf MMLB 32K} & \multicolumn{1}{c}{\bf SlideVQA} & \multicolumn{1}{c}{\bf Helmet} & \multicolumn{1}{c}{\bf LongBench v2} & \multicolumn{1}{c}{\bf DUDE} & \multicolumn{1}{c}{\bf TableVQA} \\
  \midrule
  Text & 81.6 \textcolor{teal}{(+1.1)} & 82.7 \textcolor{teal}{(+0.3)} & 42.8 \textcolor{teal}{(+2.3)} & 45.2 \textcolor{teal}{(+2.0)} & 63.5 \textcolor{red}{(-5.4)} & 74.7 \textcolor{teal}{(+4.9)} & 67.1 \textcolor{red}{(-0.3)} & 62.5 & 41.0 \textcolor{red}{(-2.0)} & 52.0 \textcolor{red}{(-0.1)} & 75.6 \textcolor{red}{(-0.8)} \\
  Visual & 80.5 & 82.4 & 40.5 & 43.2 & 68.9 & 69.8 & 67.4 & 62.6 & 43.0 & 52.1 & 76.4 \\
  \bottomrule
  \end{tabular}
  }%
  \end{center}
  \caption{A comparison of the text and visual branches of synthetic reasoning V1. We find that the text branch is slightly better than the visual branch (+1.1 VA) and significantly better on MMLongBenchDoc and MMLBD-C (+2.3 and +2.0 MMLBD-C respectively).}\label{tab:text_vs_visual_branch}
\end{table}

\section{Effect of CPT}
\label{sec:effect_of_cpt}

One might question to what extent the CPT checkpoints which we merged into our synthetic reasoning models are responsible for our results. Thus, we provide an extended version of Table~\ref{tab:main_comparisons} in Table~\ref{tab:effect_of_cpt}. We note that CPT checkpoints were also merged into the Plain Distillation models and as noted in \citep{orion}, SFT and CPT did not compose well for significantly higher VA scores. Given this, we find that the gains of synthetic reasoning are distinct from the gains of CPT and from the SFT models from \citep{orion}.

\begin{table}[t]
  \begin{center}
  \resizebox{\textwidth}{!}{%
  \begin{tabular}{lccccccccccc}
  \toprule
  \multicolumn{1}{l}{\bf Model} & \multicolumn{1}{c}{\bf VA} & \multicolumn{1}{c}{\bf LCA} & \multicolumn{1}{c}{\bf MMLBD} & \multicolumn{1}{c}{\bf MMLBD-C} & \multicolumn{1}{c}{\bf MMLB 128K} & \multicolumn{1}{c}{\bf MMLB 32K} & \multicolumn{1}{c}{\bf SlideVQA} & \multicolumn{1}{c}{\bf Helmet} & \multicolumn{1}{c}{\bf LongBench v2} & \multicolumn{1}{c}{\bf DUDE} & \multicolumn{1}{c}{\bf TableVQA} \\
  \midrule
  \multicolumn{12}{l}{\bf Qwen3 VL} \\
  ~~235B A22B Instruct & \textbf{98.4} & \textbf{98.5} & 54.8 & 56.2 & \textbf{78.6} & \textbf{82.4} & \textbf{84.5} & 67.6 & \textbf{50.0} & 59.1 & 78.8 \\
  ~~\textbf{Synthetic Reasoning (Ours)} & 95.0 \textcolor{teal}{(+1.3)} & 94.4 \textcolor{teal}{(+2.3)} & \textbf{55.8} \textcolor{teal}{(+4.0)} & \textbf{58.2} \textcolor{teal}{(+4.4)} & 75.7 \textcolor{teal}{(+5.3)} & 78.6 \textcolor{red}{(-0.3)} & 75.4 \textcolor{red}{(-1.8)} & \textbf{68.5} \textcolor{teal}{(+5.5)} & 43.0 \textcolor{teal}{(+1.0)} & 55.1 \textcolor{red}{(-6.7)} & \textbf{81.6} \textcolor{teal}{(+4.0)} \\
  ~~32B Instruct & 93.7 & 92.1 & 51.8 & 53.8 & 70.4 & 78.9 & 77.2 & 63.0 & 42.0 & \textbf{61.8} & 77.6 \\
  ~~CPT & 92.4 \textcolor{red}{(-1.3)} & 90.9 \textcolor{red}{(-1.2)} & 53.6 \textcolor{teal}{(+1.8)} & 55.9 \textcolor{teal}{(+2.1)} & 72.0 \textcolor{teal}{(+1.6)} & 78.2 \textcolor{red}{(-0.7)} & 69.7 \textcolor{red}{(-7.5)} & 64.6 \textcolor{teal}{(+1.6)} & 40.0 \textcolor{red}{(-2.0)} & 57.5 \textcolor{red}{(-4.3)} & 81.1 \textcolor{teal}{(+3.5)} \\
  \midrule
  \multicolumn{12}{l}{\bf Mistral} \\
  ~~\textbf{Synthetic Reasoning (Ours)} & 87.8 \textcolor{teal}{(+7.6)} & 87.9 \textcolor{teal}{(+11.3)} & 47.5 \textcolor{teal}{(+7.6)} & 49.3 \textcolor{teal}{(+7.9)} & 75.4 \textcolor{teal}{(+9.0)} & 75.0 \textcolor{teal}{(+2.1)} & 69.7 \textcolor{teal}{(+1.9)} & 62.6 \textcolor{teal}{(+25.6)} & 43.0 \textcolor{teal}{(+4.0)} & 54.1 \textcolor{teal}{(+1.3)} & 77.6 \textcolor{teal}{(+23.6)} \\
  ~~CPT & 84.0 \textcolor{teal}{(+3.8)} & 84.1 \textcolor{teal}{(+7.5)} & 41.5 \textcolor{teal}{(+1.6)} & 42.7 \textcolor{teal}{(+1.3)} & 69.3 \textcolor{teal}{(+2.9)} & 74.5 \textcolor{teal}{(+1.6)} & 68.0 \textcolor{teal}{(+0.2)} & 51.7 \textcolor{teal}{(+14.7)} & 47.0 \textcolor{teal}{(+8.0)} & 60.1 \textcolor{teal}{(+7.3)} & 73.4 \textcolor{teal}{(+19.4)} \\
  ~~3.1 Small 24B & 80.2 & 76.6 & 39.9 & 41.4 & 66.4 & 72.9 & 67.8 & 37.0 & 39.0 & 52.8 & 54.0 \\
  \bottomrule
  \end{tabular}
  }%
  \end{center}
  \caption{Our main checkpoints and baselines vs the CPT checkpoints, deltas relative to the base models. We find that the gains of synthetic reasoning are distinct from the gains of CPT.}\label{tab:effect_of_cpt}
\end{table}

\begin{figure}[t]
  \begin{center}
  \includegraphics[width=0.73\linewidth]{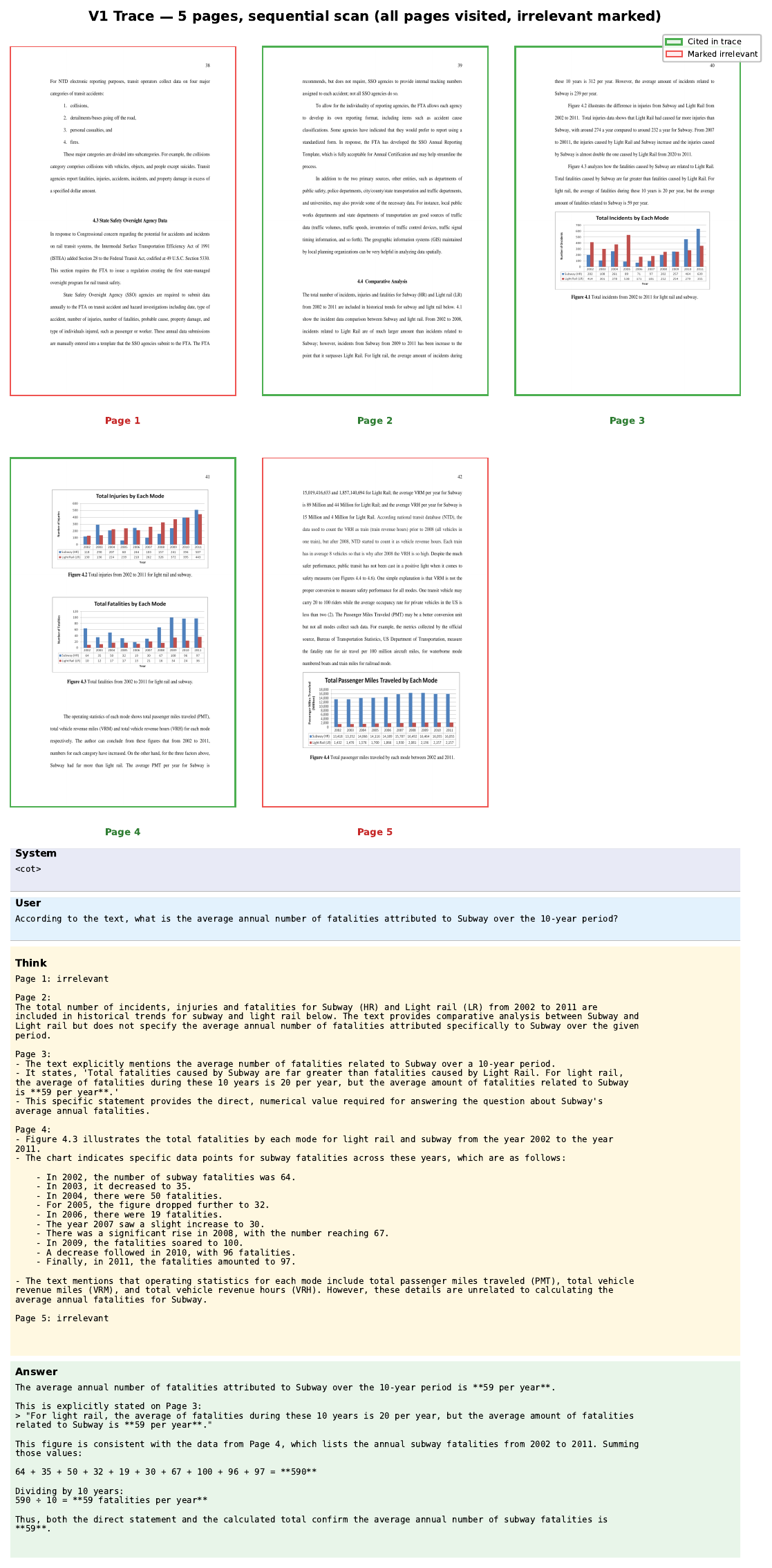}
  \end{center}
  \caption{An example from the v1 dataset.}
  \label{fig:v1_trace_example}
\end{figure}

\begin{figure}[t]
  \begin{center}
  \includegraphics[width=0.73\linewidth]{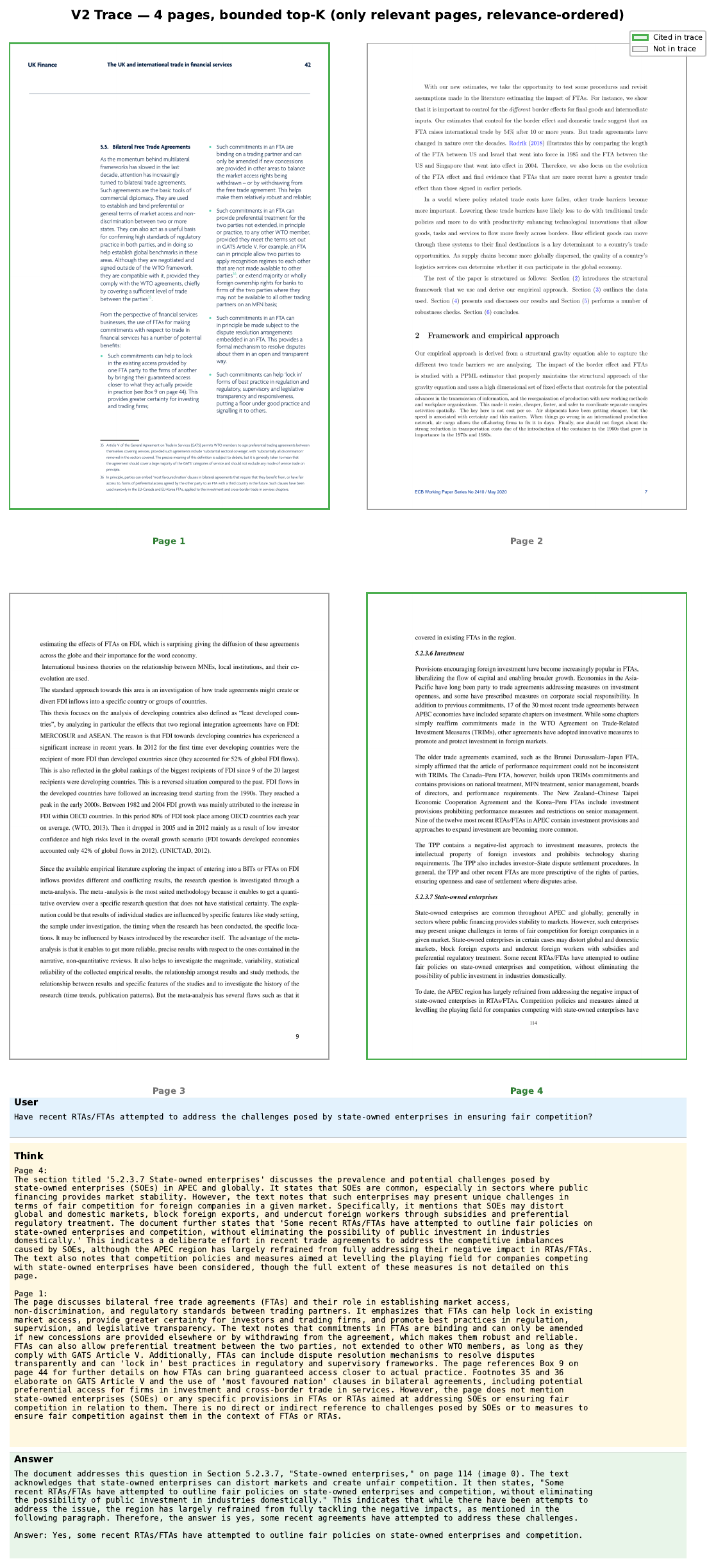}
  \end{center}
  \caption{An example from the v2 dataset.}
  \label{fig:v2_trace_example}
\end{figure}

\section{External SFT data composition}
\label{sec:external_sft_composition}

Following \citep{orion}, tables~\ref{tab:luth_composition} and~\ref{tab:smoltalk2_composition} detail the normalized composition of the external SFT data used in our experiments. We draw samples according to the distributions in the table.

\begin{table}[h]
  \centering
  \begin{tabular}{lc}
  \toprule
  \textbf{Source} & \textbf{Proportion (\%)} \\
  \midrule
  Scholar & 30.0 \\
  Smoltalk2 & 30.0 \\
  Aya Dataset & 10.0 \\
  Tulu3 Persona Math & 10.0 \\
  Tulu3 Persona Instruct & 10.0 \\
  OpenHermes & 10.0 \\
  \bottomrule
  \end{tabular}
\caption{Composition of the Luth \citep{luthsft} SFT data mixture \citep{orion}.}
\label{tab:luth_composition}
\end{table}

\begin{table}[h]
  \centering
  \begin{tabular}{lc}
  \toprule
  \textbf{Source} & \textbf{Proportion (\%)} \\
  \midrule
  LongAlign 64K & 18.0 \\
  Mixture of Thoughts (Science) & 18.0 \\
  OpenThoughts3 1.2M & 18.0 \\
  TableGPT & 18.0 \\
  Tulu3 SFT Personas Instruction Following & 9.0 \\
  Smoltalk Multilingual (8 languages) & 3.6 \\
  Smoltalk Smol Magpie Ultra & 3.6 \\
  Smoltalk Smol Summarize & 3.6 \\
  Multifaceted Collection & 3.6 \\
  OpenHermes 2.5 & 1.8 \\
  EverythingLM-data-V3 & 1.8 \\
  Smoltalk Everyday Conversations & 0.9 \\
  \bottomrule
  \end{tabular}
  \caption{Composition of the Smoltalk2 \citep{smoltalk2} SFT data mixture \citep{orion}.}
  \label{tab:smoltalk2_composition}
\end{table}

\end{document}